\documentclass[10pt,twocolumn,letterpaper]{article}

\usepackage{times}
\usepackage{epsfig}
\usepackage{graphicx}
\usepackage{amsmath}
\usepackage{amssymb}

\usepackage[pagebackref=true,breaklinks=true,letterpaper=true,colorlinks,bookmarks=false]{hyperref}

\begin{document}

\date{}
\title{Towards Large-Scale Pose-Invariant Face Recognition Using Face Defrontalization}

\author{Patrik Mesec\\
Aircash\\
{\tt\small patrik.mesec@aircash.eu}
\and
Alan Jović\\
University of Zagreb Faculty of Electrical Engineering and Computing\\
{\tt\small alan.jovic@fer.unizg.hr}
}

\maketitle

\begin{abstract}
   Face recognition under extreme head poses is a challenging task. Ideally, a face recognition system should perform well across different head poses, which is known as pose-invariant face recognition. To achieve pose invariance, current approaches rely on sophisticated methods, such as face frontalization and various facial feature extraction model architectures. However, these methods are somewhat impractical in real-life settings and are typically evaluated on small scientific datasets, such as Multi-PIE. In this work, we propose the inverse method of face frontalization, called face defrontalization, to augment the training dataset of facial feature extraction model. The method does not introduce any time overhead during the inference step. The method is composed of: 1) training an adapted face defrontalization FFWM model on a frontal-profile pairs dataset, which has been preprocessed using our proposed face alignment method; 2) training a ResNet-50 facial feature extraction model based on ArcFace loss on a raw and randomly defrontalized large-scale dataset, where defrontalization was performed with our previously trained face defrontalization model. Our method was compared with the existing approaches on four open-access datasets: LFW, AgeDB, CFP, and Multi-PIE. Defrontalization shows improved results compared to models without defrontalization, while the proposed adjustments show clear superiority over the state-of-the-art face frontalization FFWM method on three larger open-access datasets, but not on the small Multi-PIE dataset for extreme poses (75 and 90 degrees). The results suggest that at least some of the current methods may be overfitted to small datasets. 
\end{abstract}

\section{Introduction}

Face recognition is the task of identifying or verifying the identity of a person using their face from a digital image \cite{FACE-RECOGNITION-HANDBOOK}. A person can be identified by comparing their encoded face to a database of encoded faces of known people (one-to-many matching). 
Face verification is a one-to-one matching task in which a system is presented with the facial image of an unknown person along with a statement of identity (encoded face). 
Face recognition has become one of the most widely used tasks in biometric authentication systems because it can be used in many domains (e.g., surveillance, home security, border control) \cite{FACE-RECOGNITION-SURVEY, FACE-RECOGNITION-SURVEY-2, FACE-RECOGNITION-REVIEW-2020}. The main challenge in developing face recognition systems is the reliable and robust extraction of facial features (face encoding) \cite{FACE-RECOGNITION-HANDBOOK}. The main idea behind feature extraction is that more similar faces should have more similar features. Ideally, the extracted facial features should be independent of image illumination, head pose, facial expression, occlusion, and other attributes \cite{Illumination, Pose_invariant, Occlusion}. 

\begin{figure}
  \centering
  \includegraphics[width=\linewidth]{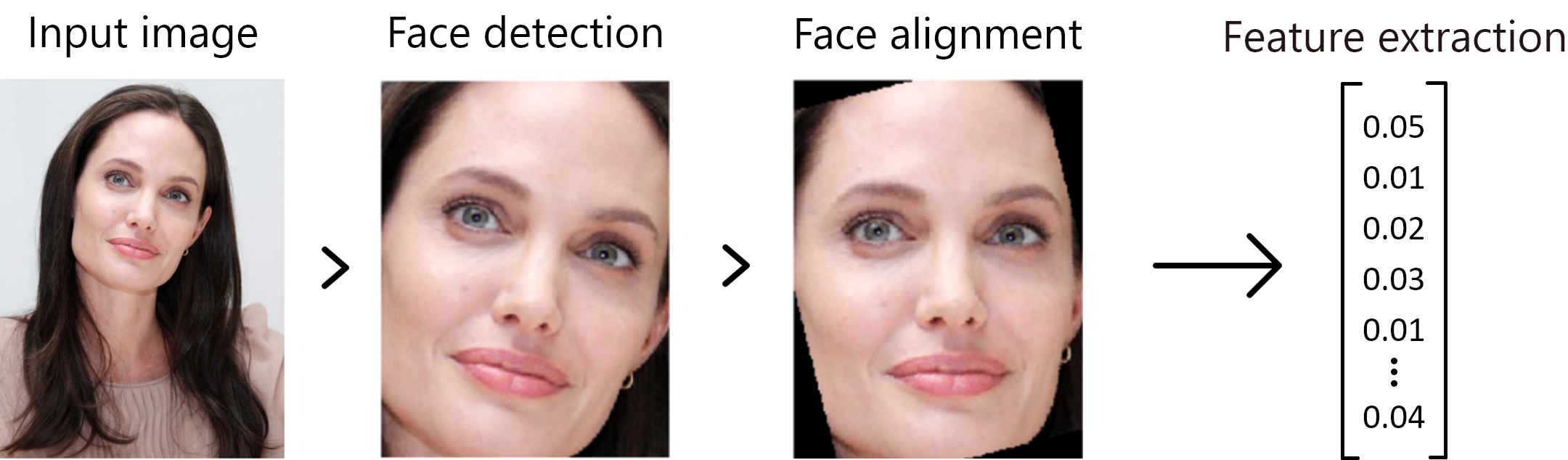}
  \caption{The main components of the face feature extraction pipeline}
  \label{fig:face-recognition-pipeline}
\end{figure}

The facial feature extraction pipeline is shown in Figure~\ref{fig:face-recognition-pipeline}. The first step of the pipeline is face detection in the input image using a face detection model. The detected face is then aligned using an affine transformation. Lastly, facial features in the form of representation vectors are extracted. The state-of-the-art in facial feature extraction is based on deep learning \cite{ARC-FACE, DEEP-FACE-RECOGNITION-SURVEY}. While face alignment reduces some variation in head pose, extreme head poses (i.e., profile or near-profile poses) remain a challenge. Currently, the best results in face recognition are obtained for frontal poses \cite{cfp-fp}, for which all components of the pipeline perform best. In this work, we investigate how to improve face recognition under extreme head poses, thus we aim to achieve pose-invariant face recognition, where the extracted facial features are invariant to the angle of the head poses. Certainly, constructing a highly accurate, pose-invariant face recognition model is of great importance for many current applications. It would improve the usability, robustness, and resilience of currently available face recognition systems. 

There are many approaches proposed to handle pose-invariant face recognition \cite{TP-GAN, SPAE, PAM-1, Pose_invariant}. The main reason why currently the best results in face recognition are obtained for frontal poses is data-related. Namely, the datasets that are used to train deep learning models need to be large and diverse to avoid overfitting and data-induced bias. A dataset should contain many identities under different conditions, including head poses, and it is quite difficult to collect so many images. There are many public face recognition datasets, but most of them mainly contain frontal head poses (e.g., LFW \cite{LFW}, WEBFACE-CASIA \cite{WEBFACE-CASIA}, IJB-B \cite{IJB-B}). On the other hand, there are datasets, such as Multi-PIE \cite{MULTI-PIE}, which contain a variety of facial attributes, including different head poses, but not many identities. 

One approach to address data acquisition issues and solve pose-invariant face recognition is an additional image preprocessing step where the head pose is normalized \cite{SECOND-TRADITIONAL}. The normalized head pose is a neutral, frontal head pose, from which facial features are extracted. The normalization requires an additional generative model that recreates the frontal face from a given head pose - the face frontalization model. In this approach, the feature extraction model does not need to be changed, but an additional face frontalization model is required in both the training and inference pipelines \cite{TP-GAN}. Generative models are often cumbersome, and therefore the inference speed of the pipeline suffers, which can be critical in real-time applications \cite{FACE-RECOGNITION-HANDBOOK}. Still, there are many face frontalization approaches and models, but a common problem is generalizing to unseen head poses \cite{FFWM}. Recently, several approaches used synthetic data \cite{Fake_it_till_ymi, SynFace} to train face recognition models. 

In this work, we propose the inverse method of face frontalization, called face defrontalization, to augment the training dataset of facial feature extraction model with the aim of improving pose-invariant face recognition. We propose to train a face defrontalization model that accepts aligned frontal faces as input and outputs aligned faces at the extreme 90 degrees head angle (i.e., the profile head poses). Face defrontalization is arguably an easier task than face frontalization, especially in extreme cases, since there is more information on the front of the face than on other head poses. With our approach, the face recognition pipeline remains the same during inference (Figure \ref{fig:face-recognition-pipeline}), and the only change occurs during training of the facial feature extraction model. Our face defrontalization model is an adaptation of the Flow-based Feature Warping Model (FFWM), which was originally used as a face frontalization model \cite{FFWM}. While our facial feature extraction model must be trained from scratch, the advantage is that the speed of the face recognition pipeline at the inference step (i.e., when recognizing a new face) is unaffected. Face alignment of the face defrontalization model training dataset is an important preprocessing step. However, due to inconsistent facial feature detection under extreme head poses, we additionally propose a novel face alignment method aimed at achieving more consistent alignment for extreme head poses. To the best of the authors' knowledge, the use of the proposed face defrontalization-based augmentation technique to train a facial feature extraction model on a large scale has never been reported. We summarize two contributions of this work as follows: 
\begin{enumerate}
    \item A face defrontalization method that augments the training step of the facial feature extraction model and enhances pose-invariant feature extraction.
    \item A preprocessing face alignment method for extreme head poses.
\end{enumerate}

Our work is organized, as follows. In Section~\ref{related_work}, we review the related work in pose-invariant face recognition. Section~\ref{methodology} describes the methods and datasets used. In Section~\ref{experiments}, we provide details of the experiments and present the obtained results. We discuss the main findings in Section~\ref{discussion}. Section~\ref{conclusion} concludes the paper.

\section{Related Work}
\label{related_work}
Face frontalization is a task in which a frontal face is generated from a given profile face. This task has been widely explored in research, especially since the invention of deep generative adversarial neural networks (GANs) \cite{GAN, TP-GAN, DA-GAN, FFWM, PIM, CFR-GAN}. Some traditional methods tackle the face frontalization problem by 2D/3D local texture warping \cite{FIRST-TRADITIONAL, SECOND-TRADITIONAL} or statistical modeling \cite{THIRD-STATISTICAL-TRADITIONAL}, but more recent approaches are based on deep learning. Huang \emph{et al.} \cite{TP-GAN} propose a two-pathway generative adversarial network (TP-GAN) for the photorealistic synthesis of frontal views by simultaneously perceiving global structures and local details. Yin \emph{et al.} \cite{DA-GAN} propose a self-attention-based generator that integrates local features with their long-range dependencies to provide better feature representations. Wei \emph{et al.} \cite{FFWM} propose an FFWM model composed of GAN and an optical flow model. The optical flow model is used to define pixel correspondence between frontal and profile faces in the feature space. Most frontal face approaches are supervised, but Qian \emph{et al.} \cite{FNM} use unsupervised learning with unpaired face images in the wild (i.e., in unconstrained environments). Some face frontalization methods are 3D-based \cite{ HF-PIM, FF-GAN, UV-GAN}. To alleviate the problem of extreme poses without generating any frontal facial image, Wang \emph{et al.} \cite{pseudo-extreme-poses} propose pseudo facial generation.

A more general approach than face frontalization is multi-view face synthesis. Kan \emph{et al.} \cite{SPAE} stack multiple shallow auto-encoders that can progressively convert non-frontal face images into frontal images, which means that pose variations are gradually narrowed down to zero. Similarly, Xu \emph{et al.} \cite{FFLOWGAN} divide large-angle face synthesis into a series of simple small-angle rotations, but with an additional flow module aimed at computing a dense correspondence between the input and target faces. Hu \emph{et al.} \cite{CAPG-GAN} propose CAPG-GAN, which generates an arbitrary face pose directly from a given face pose using heatmaps of facial landmarks as additional input. Yim \emph{et al.} \cite{REMOTE-CODE} use a special code as an additional input that encodes the target pose.

Face frontalization and multiview face synthesis are great approaches to augment a dataset or normalize facial poses to achieve pose-invariant face recognition. However, much work has been done on facial feature extraction models to extract better pose-invariant features. Cao \emph{et al.} \cite{DREAM} present the DREAM block to map the profile face pose into the frontal face pose in the feature space. The DREAM block can be easily added to an existing facial feature extraction model. Masi \emph{et al.} \cite{ PAM-1, PAM-2, PAM-3} address pose invariance through several pose-specific models. In addition, they use pose classification to forward the image to the corresponding pose-aware CNN model (PAM). Yin \emph{et al.} \cite{MTL} propose a pose-directed multi-task CNN in which different poses are grouped to learn pose-specific identity features simultaneously for all poses. Unlike traditional CNNs that use fixed convolution kernels, Xiong \emph{et al.} \cite{c-CNN} propose a conditional CNN (c-CNN) with dynamically activated kernel sets, where different kernel sets are conditionally activated for different input face poses. Zhang \emph{et al.} \cite{Pose_invariant} propose a pose-adaptive angular distillation loss to mitigate the negative effect of uneven distribution of face poses in the training dataset to pay more attention to the samples with large pose variations. He \emph{et al.} \cite{3D-Reconstruction} enhance face recognition with a bypass of self-supervised 3D reconstruction, which enforces the neural backbone to focus on the identity-related depth and albedo information.

The facial feature extraction model is the most important part of accurate face recognition. The current state-of-the-art facial feature extraction model is based on deep convolutional neural networks (DCNN) \cite{ARC-FACE}. DCNN maps the face image into features (embeddings) that should have a small intraclass distance and a large interclass distance. There are two different approaches to obtaining these features. The first approach is based on training a multi-class DCNN classifier that can separate different identities in the training set, e.g., by using a softmax classifier \cite{VGG-FACE2, DEEP-FACE-FIRST, DEEP-FACE-RECOGNITION}. The most important part of this approach is the loss function. State-of-the-art results are currently obtained with the ArcFace loss function \cite{ARC-FACE}, but there are many other loss functions explored \cite{MARGIN-loss, CENTRE-loss, COS-FACE, SPHERE-FACE, RANGE-loss, Additive-loss}. During inference, the classification head layer is removed and the final features are extracted from the previous layer. The other approach is based on the triplet loss function \cite{FACE-NET}, where the final features are learned directly.

In this work, we use and train a state-of-the-art facial feature extraction model based on ArcFace loss with InsightFace \cite{insightface} implementation. We also train a face defrontalization model based on FFWM \cite{FFWM}. The input to our defrontalization model is a frontal face, and the output is a face at an extreme pose angle. During the training of the facial feature extraction model, we defrontalize frontal faces randomly, but under control (see Section \ref{facial_feature_extraction_model}), using the trained face defrontalization model. The speed of the face recognition inference step remains unchanged. Thus, we combine the advantages of the mentioned works to achieve better pose-invariant features at a large scale.       



\section{Methodology}
\label{methodology}

\begin{figure*}
  \centering
  \includegraphics[width=0.9\linewidth]{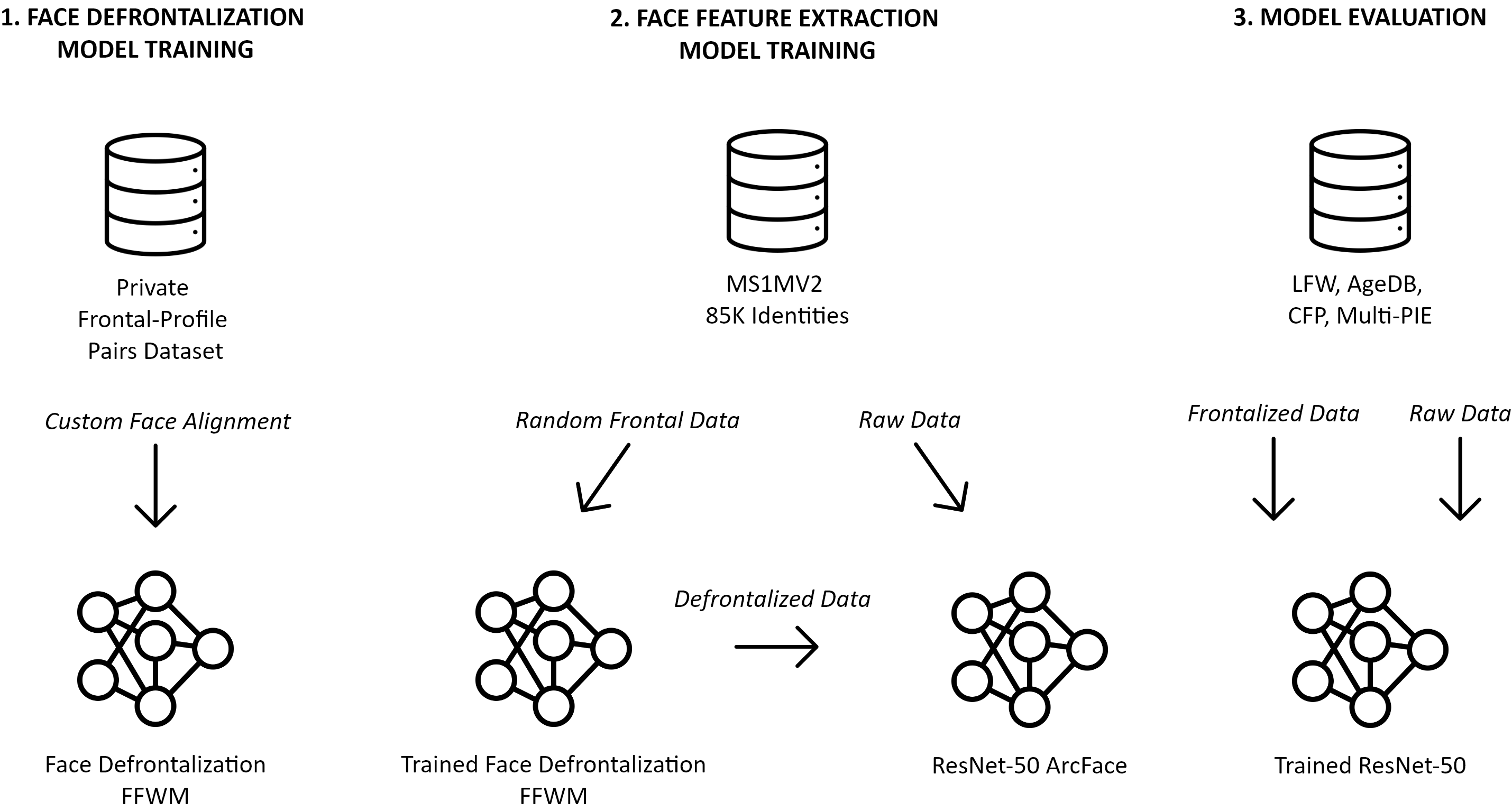}
  \caption{Proposed methodology for a large-scale pose-invariant face recognition}
  \label{fig:paper-workflow}
\end{figure*}

The overall methodology proposed in this work is depicted in Figure~\ref{fig:paper-workflow}. It consists of three steps: 

\begin{enumerate}
    \item A face defrontalization FFWM model is trained on a private frontal-profile pairs dataset and preprocessed using our proposed face alignment method.
    \item A ResNet-50 facial feature extraction model based on ArcFace loss is trained on a raw and randomly defrontalized MS1MV2 \cite{ms1m} dataset, where defrontalization was performed with our trained face defrontalization model from the previous step. 
    \item The trained ResNet-50 facial feature extraction model is evaluated on frontalized data with the publicly available FFWM \cite{FFWM} model and raw data from LFW \cite{LFW}, AgeDB \cite{agedb}, CFP \cite{cfp-fp}, and Multi-PIE \cite{MULTI-PIE} datasets.
\end{enumerate}

In the following subsections, the first two steps of the methodology are described in detail.

\subsection{Face alignment}
Face alignment precedes the face defrontalization model. It is important to properly align and preprocess the input facial images. The target profile images should also be aligned consistently. This ensures that the model is able to effectively learn to synthesize profile images from frontal images. In addition, the generated output profile face image is aligned and can be used to extract facial features without other processing. Frontal input face image is aligned using the ArcFace \cite{ARC-FACE} alignment. Dense facial landmarks are detected using Face++ \cite{faceplusplus}. For the target profile face image, we use a different method to estimate the ArcFace alignment. Namely, the problem with aligning profile face images at extreme angles is inaccurate and inconsistent detection of facial landmarks, resulting in incorrect face alignment. To alleviate this, we use several facial landmarks and an aligned frontal image of the same identity as a reference for profile face image alignment. The alignment of a pair of frontal and profile face images of the same identity is performed with the following procedure (Figure \ref{fig:face-alignment}):

\begin{figure}
  \centering
  \includegraphics[width=0.8\linewidth]{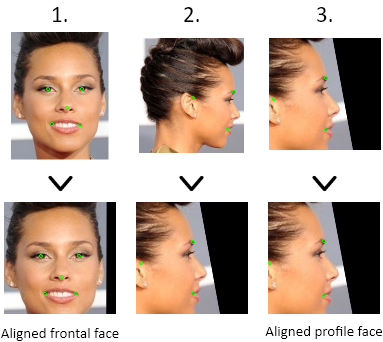}
  \caption{Visualization of our proposed face alignment steps}
  \label{fig:face-alignment}
\end{figure}

\begin{enumerate}
  \item The frontal face image is aligned with the ArcFace alignment method, which uses five landmarks on the face: the left and right eyes, the top of the nose, and the left and right corners of the mouth.
  \item Three landmarks are used to align the profile face: the top of the nose midline, the corner of the mouth (depending on the side of the face), and a point near the ear. First, additional information from the aligned frontal face image is used, namely, the exact y-coordinate positions of the top of the nose midline and one of the corners of the mouth of the aligned frontal face image. The transformation of the profile face image is performed in such a way that the y-coordinates of these two landmarks are positioned as they were obtained from the frontal face image, and the x-coordinates are set to half the image size. This ensures that the scale between the profile and frontal face images is identical and that the orientation between the aligned profile face images is more consistent than when using the ArcFace method directly, which often fails due to poor recognition of the landmarks at extreme facial poses. Through visualization, it was found that, in most cases, the landmarks of the top of the nose midline and the corners of the mouth are orthogonal to each other, which is the reason for setting the x-coordinates to the same value at half the width of the aligned image.
  \item After the transformation, which ensures the correct positioning of the two landmarks on the profile face, the remaining landmark is located near the ear. The reason for using this landmark is to ensure a consistent display of the aligned profile face image where the ear is not present. Therefore, we set the x-coordinate of this landmark to 0. The reason this step is separate from the previous one is that we cannot accurately determine the y-coordinate destination of this landmark while ensuring the correct position of the other two landmarks. Therefore, we set the y-coordinate of this landmark to the transformed value from the previous step and perform the final transformation. 
\end{enumerate}

\subsection{Face defrontalization model}
\begin{figure}
  \centering
  \includegraphics[width=\linewidth]{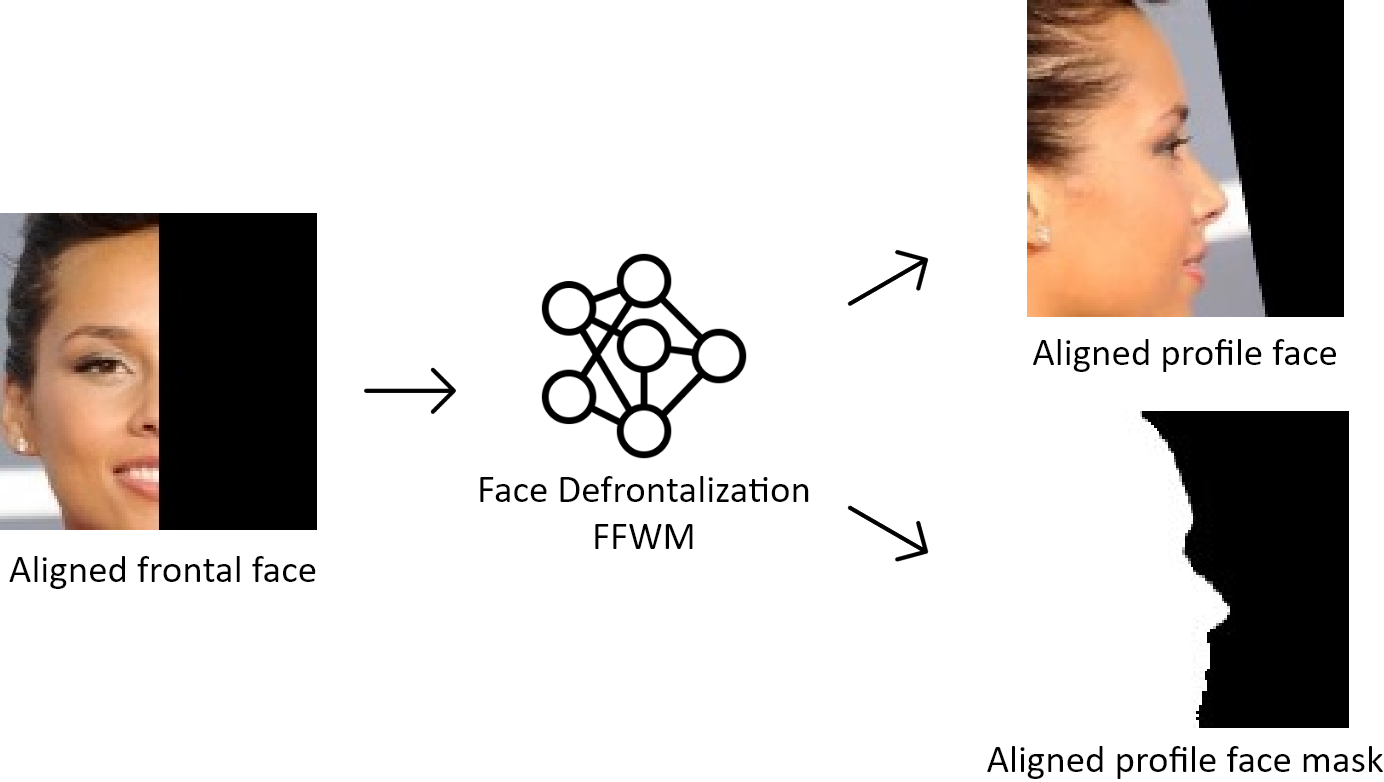}
  \caption{Input and outputs visualization of our face defrontalization model}
  \label{fig:ffwm-input-output}
\end{figure}

We developed a face defrontalization model based on FFWM \cite{FFWM}. FFWM consists of four models: forward optical flow network, backward optical flow network, generator network, and discriminator network. The input to our modified model is a frontal, aligned, and horizontally bisected face image of size 112x112, and the output is a profile face image in the extreme pose of the same size (Figure~\ref{fig:ffwm-input-output}). 

The image size is chosen to match the input size of the facial feature extraction model (see Section~\ref{facial_feature_extraction_model}), so we do not need to do any additional processing on the images during the training of the feature extraction model. The use of image bisection allows one to choose which side of the face to defrontalize, and that is important because we can obtain two profile pictures from one frontal face. To modify the FFWM model for our face defrontalization task, we removed one encoding and one decoding layer of the optical flow models because of the changed input image size (from 128x128 to 112x112), resulting in optical flow models with fewer parameters. Specifically, our optical flow models have 7M parameters, while the original optical flow models have 52M parameters. We also added an output layer to each decoder layer of the generator network to predict the mask of the synthesized profile face on multiple scales. In this way, we can change the background of the synthesized profile face.

The FFWM model uses a combination of techniques to synthesize photorealistic and illumination-preserving profile images from frontal images. These techniques include the use of optical flow to estimate the motion of pixels in the frontal image as they are transformed into the corresponding pixels in the profile image, warping layers and attention mechanisms to align the frontal and profile images, and a GAN to improve the photorealism of the synthesized images. The model also includes an Illumination Preserving Module (IPM) responsible for learning how to synthesize illumination-preserving images from inconsistent image pairs. IPM includes two pathways that collaborate to ensure the synthesized profile images are illumination-preserving and with fine details. 

Our face defrontalization FFWM model uses several loss functions for the training process. These include:

\textbf{Pixel-wise Loss}: 
\begin{equation}
L_{pixel}=\sum_{s=1}^{S} ||\hat{I}^{synth}_s-I^{gt}_s||_1
\end{equation}

where $I^{gt}_{s}$ is the ground truth profile image, $\hat{I}^{synth}_{s}$ is the synthesized profile image, and $s$ is the scale. We use three scales, which are 28x28, 56x56, and 112x112.

\textbf{Perceptual Loss}: 
\begin{equation}
 L_{p} = \sum_{i} w^i || \phi{(\hat{I}^{synth})} - \phi{({I}^{gt})} ||_1 
\end{equation}

where $\phi{(\cdot)}_i$ denotes the output of the $i$-th VGG-19 \cite{VGG-19} layer. We follow the original FFWM setting and use Conv1-1, Conv2-1, Conv3-1, Conv4-1 and Conv5-1 layer, and set $w = (1, 1/2, 1/4, 1/4, 1/8)$.

\textbf{Adversarial Loss}: 
\begin{equation}
L_{adv} = \mathbb{E}_{I^{gt}} [\log D({I^{gt}})] + \mathbb{E}_{\hat{I}^{synth}} [\log (1 - D(\hat{I}^{synth}))]
\end{equation}
Here, $\hat{I}^{synth}$ is synthesized by the generator model and $D$ is the discriminator model. 

\textbf{Illumination Preserving Loss}:
\begin{equation}
 L_{ip} = \sum_{s=1}^{s} || \hat{I^w_{s}} - I_s ||_1 
\end{equation}

where $\hat{I^w_{s}}$ is the frontal warped face image from the synthesized profile face image,  $I_s$ is the input frontal face image, and $s$ is the scale with the same setting as for the pixel-wise loss.

\textbf{Identity Preserving Loss}: 
\begin{equation}
\begin{split}
L_{id} = & || \psi_{fc2}{(\hat{I^{synth}})} - \psi_{fc2}{(I^{gt})} ||_1 +\\ 
& || \psi_{pool}{(\hat{I^{synth}})} - \psi_{pool}{(I^{gt})} ||_1
\end{split}    
\end{equation}

where $\psi(\cdot)$ denotes the pretrained LightCNN-29 \cite{LightCNN-29}. $fc2$ and $pool$ denote the fully connected layer and the last pooling layer, respectively.

\textbf{Mask Loss}:
\begin{equation}
 L_{mask} = \sum_{s=1}^{S} 
 BCE(\hat{M}^{synth}_s, M^{gt}_s)
\end{equation}

where $BCE$ denotes the binary cross-entropy loss, $\hat{M}^{synth}_s$ the synthesized profile face mask, $M^{gt}_s$ the ground truth profile face mask, and $s$ is the scale with the same setting as for the pixel-wise loss.  

The overall loss function for our adapted face defrontalization FFWM model can be expressed as:
\begin{equation}
L = \lambda_{0} L_{pixel} + \lambda_{1} L_{p} + \lambda_{2} L_{adv} + \lambda_{3} L_{ip}  + \lambda_{4} L_{id}  + \lambda_{5} L_{mask}
\end{equation}

We use $\lambda$ weights from the original FFWM setting, and additionally set $\lambda_{5}$ to 1, because we added $L_{mask}$.

\subsection{Facial feature extraction model training}
\label{facial_feature_extraction_model}
We use InsightFace \cite{insightface} repository to train the facial feature extraction model. We use ResNet-50 \cite{resnet} architectural model to extract facial features. The loss function used is Additive Angular Margin Loss (ArcFace) \cite{ARC-FACE}. ArcFace addresses the problem of large intra-class variations and small inter-class variations. It aims to increase the inter-class variations by adding an additive angular margin to the target logits. This leads to a larger decision boundary between classes, which helps to improve recognition performance.

During the training process, we randomly apply the trained face defrontalization model to frontal face images. Specifically, we randomly defrontalize the left or right part of the face. In addition, not all faces in the training dataset are defrontalized, but only those that are more frontal according to the face alignment error. The alignment error is calculated for all images before training according to ArcFace alignment \cite{ARC-FACE}, and we randomly defrontalize only images whose alignment error is below the predefined threshold because our face defrontalization model works best on frontal faces. We have set the alignment error threshold so that roughly 20 percent of the images in the training dataset are defrontalized. This approach helps to improve the pose invariance of the model's extracted facial features by exposing it to a wider range of facial poses during training.

\subsection{Datasets}
We train our face defrontalization model on a dataset consisting of pairs of profile and frontal face images by identity. The dataset contains about 50,000 identities. Unfortunately, due to privacy restrictions, this dataset is confidential. However, for reproducibility reasons, we expose the trained defrontalization model, which can be found in the Supplementary materials of this paper.

To train the facial feature extraction model, we use the MS1MV2 \cite{ms1m,ARC-FACE} dataset, which contains about 85,000 identities and a total of 5.8 million images. The facial feature extraction model is evaluated on four different datasets: LFW \cite{LFW}, AgeDB \cite{agedb}, CFP \cite{cfp-fp}, and Multi-PIE \cite{MULTI-PIE}.  

The LFW (Labeled Faces in the Wild) \cite{LFW} dataset is a collection of 13,233 images of 5,749 individuals, all from the internet. It contains 6,000 testing pairs. We also use Cross-Age and Cross-Pose LFW testing pairs variants, namely CALFW \cite{CALFW} and CPLFW \cite{CPLFW} respectively, which both contain 6000 testing pairs. The AgeDB \cite{agedb} dataset was created to evaluate the performance of age estimation models. It contains 16,516 images of 570 individuals collected from the internet. We use the testing protocol with at least a 30-year age difference between pairs of faces, called AgeDB-30, which contains 6,000 testing pairs. CFP \cite{cfp-fp} is a dataset that aims to isolate the factor of pose variation with respect to extreme poses such as profile views. It contains 10 frontal and 4 profile images of 500 individuals, resulting in a total of 7,000 images. We use both frontal-to-profile and frontal-to-frontal testing pairs, namely CFP-FP and CFP-FF respectively, which both contain 7,000 testing pairs. The Multi-PIE \cite{MULTI-PIE} contains over 750,000 images of 337 individuals captured in 15 poses and under 19 different illumination conditions. The images were captured in a controlled environment. We test our model on Setting 2 which consists of 137 identities \cite{MULTI-PIE}. The Multi-PIE dataset allows us to evaluate faces in different poses in more detail. 

In Table \ref{table1}, we show the mean absolute differences in the estimated face pose angles of the LFW, AgeDB-30, and CFP-FP test pairs. Yaw, roll, and pitch refer to head rotations around the vertical, sagittal, and transverse axes, respectively \cite{YawRollPitch}. With these datasets, we can effectively evaluate our models on frontal and profile face images. 

\begin{table}
\begin{center}
\begin{tabular}{|c|c|c|c|}
\hline
Dataset & Pitch & Yaw & Roll\\
\hline \hline
LFW & 12.21 & 28.09 & 15.41 \\
AgeDB-30 & 11.58 & 30.51 & 18.86 \\
CFP-FP & 12.40 & 68.67 & 19.78\\
\hline
\end{tabular}
\end{center}
\caption{Mean absolute differences in the estimated face pose angles of the test pairs for the three considered datasets}
\label{table1}
\end{table}



\section{Experiments}
\label{experiments}
\subsection{Experimental setting}
The face defrontalization model was trained based on the adjusted FFWM model. The training was performed with similar hyperparameter settings as for the original FFWM frontalization model \cite{FFWM}. The main difference was in the number of epochs required for the generator model to converge on the private dataset. Our private dataset has about 50,000 identities, which is two orders of magnitude more identities than Multi-PIE, and as a result, we trained the generator model for 50 epochs instead of the original FFWM's 200 epochs. We used the same scheme for pre-training the forward and backward optical flow models, even though our optical flow models have fewer parameters. 

Training of the facial feature extraction model was performed using the InsightFace repository and its configuration for training the ResNet-50 model on the MS1MV2 dataset. The configuration included an initial learning rate of 0.1 with polynomial decay to 0, SGD optimizer with weight decay of 5e-4 and momentum of 0.9, and 20 epochs. An NVIDIA GeForce GTX 2070 Super 8 GB GPU was available for the experiments, so the training was carried out by accumulating gradients to align with the predefined InsightFace's training configuration, which defines the batch size of 128 on 8 parallel graphics units. We used the batch size of 32 and a 32-step gradient accumulation. 

We trained two ResNet-50 models for facial feature extraction, with the aim of comparing the models with and without face defrontalization augmentation. In the first experiment, training without augmentation was replicated from the official InsightFace repository, and in the second experiment, we used face defrontalization augmentation. 

We compared our method with the publicly available pre-trained state-of-the-art face frontalization FFWM model on the LFW, CFP-FP, and AgeDB-30 testing pairs. We first frontalized all the LFW, CFP-FP, and AgeDB-30 pairs, and then extracted facial features with both trained ResNet-50 models (with and without defrontalization). Additionally, we compared trained ResNet-50 models with and without our defrontalization method on CFP-FF, CALFW, and CPLFW testing pairs. For the evaluation of the models on all six datasets, the calculation was performed using cross-validation face pair verification accuracy on 10 folds. The final evaluation value was the average verification accuracy over 10 folds.

To analyze the performance of our face recognition model at different face poses and compare it with the other related methods, the trained models were also evaluated on the Multi-PIE dataset, which contains several face poses under controlled conditions. The evaluation was performed in accordance with Setting 2 of Multi-PIE, which contains 137 identities \cite{MULTI-PIE}, and the top-1 accuracy was measured. 

The code for using our method, along with the experimental settings mentioned here, is available in the Supplementary materials of this work.

\subsection{Results}
\begin{figure}
  \centering
  \includegraphics[width=\linewidth]{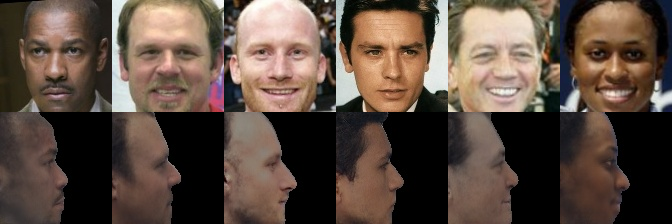}
  \caption{Qualitative results of the proposed face defrontalization model from the LFW and AgeDB datasets}
  \label{fig:face-defrontalization-predictions}
\end{figure}

The qualitative results of the trained face defrontalization model are shown in Figure~\ref{fig:face-defrontalization-predictions}. The results indicate that the model performs well in preserving eyebrow and nose details, but worse in retaining the illumination of the image, which could be due to the smaller number of parameters of the optical flow models compared to the original ones.   

\begin{table*}[!ht]
\begin{center}
\begin{tabular}{|c|c|c|c|c|c|c|}
\hline
Method & LFW & AgeDB-30 & CFP-FP & CFP-FF & CALFW & CPLFW\\
\hline \hline
FFWM, ResNet-50 Baseline & 99.517 & 96 & 85.15 & - & - & -\\
FFWM, ResNet-50 \& defrontalization & 99.56 & 95.4 & 85.81 & - & - & -\\
ResNet-50 Baseline & 99.8 & 97.917 & 97.9 & 99.8 & 96.01 & 92.66\\
ResNet-50 \& defrontalization (ours) & 99.783 & 98.017 & 98.114 & 99.829 & 96.08 & 92.967\\
\hline
\end{tabular}
\end{center}
\caption{Face verification accuracy results for the considered models, in \%}
\label{table2}
\end{table*}

The face verification accuracy results on the LFW, AgeDB-30, CFP-FP, CFP-FF, CALFW, and CPLFW testing pairs for all methods are shown in Table \ref{table2}. The accuracy of the baseline model without defrontalization is slightly higher than our model with defrontalization only on the LFW pairs, while our model achieves better results on all other testing pairs. It can be observed that the biggest improvement is on the CFP-FP and CPLFW pairs, which are frontal-to-profile datasets. We note that the models with FFWM frontalization perform worse than our models in both experiments for all tested pairs. The goal of defrontalization was to improve the performance of the model on profile faces while retaining most of the performance on frontal faces, and the obtained results confirm this.

\begin{table*}[!ht]
\begin{center}
\begin{tabular}{|c|c|c|c|c|c|c|c|}
\hline
Method & $\pm15^{\circ}$ & $\pm30^{\circ}$ & $\pm45^{\circ}$ & $\pm60^{\circ}$ & $\pm75^{\circ}$ & $\pm90^{\circ}$ & Avg \\
\hline \hline
Light CNN \cite{LightCNN-29} & 98.59 & 97.38 & 92.13 & 62.09 & 24.18 & 5.51 & 63.31 \\
DR-GAN \cite{DR-GAN} & 94.90 & 91.10 & 87.20 & 84.60 & - & - & 89.45 \\
FF-GAN \cite{FF-GAN} & 94.60 & 92.50 & 89.70 & 85.20 & 77.20 & 61.20 & 83.40 \\
TP-GAN \cite{TP-GAN} & 98.68 & 98.06 & 95.38 & 87.72 & 77.43 & 64.64 & 86.99 \\
CAPG-GAN \cite{CAPG-GAN} & 99.82 & 99.56 & 97.33 & 90.63 & 83.05 & 66.05 & 89.41 \\
PIM \cite{PIM} & 99.30 & 99.00 & 98.50 & 98.10 & 95.00 & 86.50 & 96.07 \\
3D-PIM \cite{3D-PIM} & 99.64 & 99.48 & 98.81 & 98.37 & 95.21 & 86.73 & 96.37 \\
DA-GAN \cite{DA-GAN} & 99.98 & 99.88 & 99.15 & 97.27 & 93.24 & 81.56 & 95.18 \\
HF-PIM \cite{HF-PIM} & 99.99 & 99.98 & 99.98 & 99.14 & 96.40 & 92.32 & 97.97 \\
FFWM \cite{FFWM} & 99.86 & 99.80 & 99.37 & 98.85 & 97.20 & 93.17 & 98.04 \\
ResNet-50 Baseline  & 100 & 100 & 100 & 99.5 & 95.64 & 70.03 & 94.25 \\
ResNet-50 \& defrontalization (ours) & 100 & 100 & 100 & 99.83 & 97 & 78.6 & 95.9 \\
\hline
\end{tabular}
\end{center}
\caption{Top-1 recognition rates (\%) across poses under Setting 2 of the Multi-PIE dataset}
\label{table3}
\end{table*}

The top-1 recognition rates of methods from related work using a face frontalization model and a LightCNN model for facial feature extraction, as well as our model with and without defrontalization on the Multi-PIE dataset under Setting 2, are shown in Table \ref{table3}. First, it can be observed that our model with defrontalization achieves equal or better results than the model without defrontalization, especially for the more extreme head poses, i.e., with angles of 60, 75, and 90 degrees. During the learning process, the model with defrontalization produces faces at an angle of 90 degrees. Improved results for the poses below 90 degrees are evidence of good generalization capabilities of the feature extraction model, effectively achieving better pose invariance of the extracted features. Secondly, among the results of other models, only a few models that use a separate model for face frontalization achieve better results than our models for extreme facial poses (75 and 90 degrees). One possible reason for this is that these face frontalization models may be overfitted on the Multi-PIE dataset because they were trained on a subset of the Multi-PIE dataset, which is limited in the number of included individuals (Multi-PIE defines a training set of 200 different identities across many poses and illuminations). On the contrary, both ResNet-50 baseline and  ResNet-50 with defrontalization models have not used Multi-PIE in the training set, which may be the cause for inferior results to some methods on the extreme head poses. An argument for other face frontalization models' overfitting on the MultiPIE dataset can be given from the results shown in Table \ref{table2}, where the FFWM model achieves significantly worse results on the CFP-FP dataset compared to our model that does not use frontalization. To confirm the finding, in Table \ref{table1} it can be observed that the CFP-FP dataset contains test pairs for which the average difference in the yaw rotation is equal to 68.67, corresponding to a comparison between a frontal pose and a pose with almost an extreme angle of the face.

\section{Discussion}
\label{discussion}
Promising results have been obtained for improving pose invariance in large-scale face recognition. However, we consider that the results obtained are still not perfect and that they can be further improved, for example by training a higher-quality model for face defrontalization. We consider that the main contribution of this work is in directing the scientific community toward achieving pose-invariant face recognition in the real world, where various constraints need to be satisfied. One such constraint is the number of face recognition requests per second that are sent to the model. Table \ref{table4} shows the inference speed of the facial feature extraction model ResNet-50 and FFWM model for face frontalization. To improve pose-invariance, we can observe that face frontalization introduces inference time overhead, while our approach that utilizes only ResNet-50 at inference does not. Furthermore, the rather poor generalization of the face frontalization model FFWM on an unseen dataset (i.e., CFP-FP, see Table \ref{table2}) further complicates the use of such models in real-world situations, due to various pose variations, facial expressions, lighting conditions, and other attributes, especially in extreme cases. The overfitting of the face frontalization models on the Multi-PIE dataset, as evidenced from the results shown in Tables \ref{table1}-\ref{table3} suggests that there is a need for a publicly available dataset with a greater number of different identities with different head poses.   

\begin{table}
\begin{center}
\begin{tabular}{|c|c|c|c|}
\hline
Model & Inference speed\\
\hline \hline
ResNet-50 & 5 ms \\
FFWM + ResNet-50 & 11 ms\\
\hline
\end{tabular}
\end{center}
\caption{Inference speeds of the models with default image input parameters}
\label{table4}
\end{table}

\section{Conclusion}
\label{conclusion}
In this work, we have shown that face defrontalization with carefully performed face alignment is a valid approach for improving the face recognition pipeline. The proposed model achieves high results on several large and openly available face recognition datasets, as well as comparable or better results to related work on the smaller Multi-PIE dataset under most head poses. Successful pose-invariant face recognition on a large scale in the wild is vital for many application scenarios. Therefore, the use of our approach in the currently available face recognition systems is expected to lead to improved identification and verification results without adding any additional time overhead during the inference step. Future work to improve the model results should focus on training a more robust face defrontalization model to synthesize more realistic profile faces, and thus increase the quality of data augmentation for the facial feature extraction model. Additionally, our face defrontalization augmentation could be further explored on different facial feature extraction models.  


{\small
\bibliographystyle{ieee_fullname}
\bibliography{egbib}
}

\end{document}